# Evolution Strategies as a Scalable Alternative to Reinforcement Learning

**Tim Salimans** **Jonathan Ho** **Xi Chen** **Szymon Sidor** **Ilya Sutskever**
OpenAI

## Abstract

We explore the use of Evolution Strategies (ES), a class of black box optimization algorithms, as an alternative to popular MDP-based RL techniques such as Q-learning and Policy Gradients. Experiments on MuJoCo and Atari show that ES is a viable solution strategy that scales extremely well with the number of CPUs available: By using a novel communication strategy based on common random numbers, our ES implementation only needs to communicate scalars, making it possible to scale to over a thousand parallel workers. This allows us to solve 3D humanoid walking in 10 minutes and obtain competitive results on most Atari games after one hour of training. In addition, we highlight several advantages of ES as a black box optimization technique: it is invariant to action frequency and delayed rewards, tolerant of extremely long horizons, and does not need temporal discounting or value function approximation.

## 1 Introduction

Developing agents that can accomplish challenging tasks in complex, uncertain environments is a key goal of artificial intelligence. Recently, the most popular paradigm for analyzing such problems has been using a class of reinforcement learning (RL) algorithms based on the Markov Decision Process (MDP) formalism and the concept of value functions. Successes of this approach include systems that learn to play Atari from pixels [Mnih et al., 2015], perform helicopter aerobatics Ng et al. [2006], or play expert-level Go [Silver et al., 2016].

An alternative approach to solving RL problems is using black-box optimization. This approach is known as direct policy search [Schmidhuber and Zhao, 1998], or neuro-evolution [Risi and Togelius, 2015], when applied to neural networks. In this paper, we study Evolution Strategies (ES) [Rechenberg and Eigen, 1973], a particular set of optimization algorithms in this class. We show that ES can reliably train neural network policies, in a fashion well suited to be scaled up to modern distributed computer systems, for controlling robots in the MuJoCo physics simulator [Todorov et al., 2012] and playing Atari games with pixel inputs [Mnih et al., 2015]. Our key findings are as follows:

1. We found that the use of virtual batch normalization [Salimans et al., 2016] and other reparameterizations of the neural network policy (section 2.2) greatly improve the reliability of evolution strategies. Without these methods ES proved brittle in our experiments, but with these reparameterizations we achieved strong results over a wide variety of environments.
2. We found the evolution strategies method to be highly parallelizable: by introducing a novel communication strategy based on common random numbers, we are able to achieve linear speedups in run time even when using over a thousand workers. In particular, using 1,440 workers, we have been able to solve the MuJoCo 3D humanoid task in under 10 minutes.
3. The data efficiency of evolution strategies was surprisingly good: we were able to match the final performance of A3C [Mnih et al., 2016] on most Atari environments while using between 3x and 10x as much data. The slight decrease in data efficiency is partly offset by a reduction in required computation of roughly 3x due to not performing backpropagation and not having a value function. Our 1-hour ES results require about the same amount of computation as the published 1-day results for A3C, while performing better on 23 games tested, and worse on 28. On MuJoCo tasks, we were able to match the learned policy performance of Trust Region Policy Optimization [TRPO; Schulman et al., 2015], using no more than 10x as much data.

4. We found that ES exhibited better exploration behaviour than policy gradient methods like TRPO: on the MuJoCo humanoid task, ES has been able to learn a very wide variety of gaits (such as walking sideways or walking backwards). These unusual gaits are never observed with TRPO, which suggests a qualitatively different exploration behavior.
5. We found the evolution strategies method to be robust: we achieved the aforementioned results using fixed hyperparameters for all the Atari environments, and a different set of fixed hyperparameters for all MuJoCo environments (with the exception of one binary hyperparameter, which has not been held constant between the different MuJoCo environments).

Black-box optimization methods have several highly attractive properties: indifference to the distribution of rewards (sparse or dense), no need for backpropagating gradients, and tolerance of potentially arbitrarily long time horizons. However, they are perceived as less effective at solving hard RL problems compared to techniques like Q-learning and policy gradients. The contribution of this work, which we hope will renew interest in this class of methods and lead to new useful applications, is a demonstration that evolution strategies can be competitive with competing RL algorithms on the hardest environments studied by the deep RL community today, and that this approach can scale to many more parallel workers.

## 2 Evolution Strategies

Evolution Strategies (ES) is a class of black box optimization algorithms [Rechenberg and Eigen, 1973, Schwefel, 1977] that are heuristic search procedures inspired by natural evolution: At every iteration ("generation"), a population of parameter vectors ("genotypes") is perturbed ("mutated") and their objective function value ("fitness") is evaluated. The highest scoring parameter vectors are then recombined to form the population for the next generation, and this procedure is iterated until the objective is fully optimized. Algorithms in this class differ in how they represent the population and how they perform mutation and recombination. The most widely known member of the ES class is the covariance matrix adaptation evolution strategy [CMA-ES; Hansen and Ostermeier, 2001], which represents the population by a full-covariance multivariate Gaussian. CMA-ES has been extremely successful in solving optimization problems in low to medium dimension.

The version of ES we use in this work belongs to the class of natural evolution strategies (NES) [Wierstra et al., 2008, 2014, Yi et al., 2009, Sun et al., 2009, Glasmachers et al., 2010a,b, Schaul et al., 2011] and is closely related to the work of Sehnke et al. [2010]. Let $F$ denote the objective function acting on parameters $\theta$. NES algorithms represent the population with a distribution over parameters $p_\psi(\theta)$—itself parameterized by $\psi$—and proceed to maximize the average objective value $\mathbb{E}_{\theta\sim p_\psi} F(\theta)$ over the population by searching for $\psi$ with stochastic gradient ascent. Specifically, using the score function estimator for $\nabla_\psi \mathbb{E}_{\theta\sim p_\psi} F(\theta)$ in a fashion similar to REINFORCE [Williams, 1992], NES algorithms take gradient steps on $\psi$ with the following estimator:

$$\nabla_\psi \mathbb{E}_{\theta\sim p_\psi} F(\theta) = \mathbb{E}_{\theta\sim p_\psi} \left\{ F(\theta) \nabla_\psi \log p_\psi(\theta) \right\}$$

For the special case where $p_\psi$ is factored Gaussian (as in this work), the resulting gradient estimator is also known as *simultaneous perturbation stochastic approximation* [Spall, 1992], *parameter-exploring policy gradients* [Sehnke et al., 2010], or *zero-order gradient estimation* [Nesterov and Spokoiny, 2011].

In this work, we focus on RL problems, so $F(\cdot)$ will be the stochastic return provided by an environment, and $\theta$ will be the parameters of a deterministic or stochastic policy $\pi_\theta$ describing an agent acting in that environment, controlled by either discrete or continuous actions. Much of the innovation in RL algorithms is focused on coping with the lack of access to or existence of derivatives of the environment or policy. Such non-smoothness can be addressed with ES as follows. We instantiate the population distribution $p_\psi$ as an isotropic multivariate Gaussian with mean $\psi$ and fixed covariance $\sigma^2 I$, allowing us to write $\mathbb{E}_{\theta\sim p_\psi} F(\theta)$ in terms of a mean parameter vector $\theta$ directly: we set $\mathbb{E}_{\theta \sim p_\psi} F(\theta) = \mathbb{E}_{\epsilon \sim N(0,I)}\, F(\theta + \sigma\epsilon)$. With this setup, our stochastic objective can be viewed as a Gaussian-blurred version of the original objective $F$, free of non-smoothness introduced by the environment or potentially discrete actions taken by the policy. Further discussion on how ES and policy gradient methods cope with non-smoothness can be found in section 3.

With our objective defined in terms of $\theta$, we optimize over $\theta$ directly using stochastic gradient ascent with the score function estimator:

$$\nabla_\theta\, \mathbb{E}_{\epsilon \sim N(0,I)}\, F(\theta + \sigma\epsilon) = \frac{1}{\sigma}\, \mathbb{E}_{\epsilon \sim N(0,I)} \left\{ F(\theta + \sigma\epsilon)\, \epsilon \right\}$$

which can be approximated with samples. The resulting algorithm (1) repeatedly executes two phases: 1) Stochastically perturbing the parameters of the policy and evaluating the resulting parameters by running an episode in the environment, and 2) Combining the results of these episodes, calculating a stochastic gradient estimate, and updating the parameters.

**Algorithm 1** Evolution Strategies

**Input:** Learning rate $\alpha$, noise standard deviation $\sigma$, initial policy parameters $\theta_0$
**for** $t = 0, 1, 2, \ldots$ **do**
  Sample $\epsilon_1, \ldots \epsilon_n \sim \mathcal{N}(0, I)$
  Compute returns $F_i = F(\theta_t + \sigma\epsilon_i)$ for $i = 1, \ldots, n$
  Set $\theta_{t+1} \leftarrow \theta_t + \alpha \frac{1}{n\sigma} \sum_{i=1}^{n} F_i \epsilon_i$
**end for**

### 2.1 Scaling and parallelizing ES

ES is well suited to be scaled up to many parallel workers: 1) It operates on complete episodes, thereby requiring only infrequent communication between workers. 2) The only information obtained by each worker is the scalar return of an episode: if we synchronize random seeds between workers before optimization, each worker knows what perturbations the other workers used, so each worker only needs to communicate a single scalar to and from each other worker to agree on a parameter update. ES thus requires extremely low bandwidth, in sharp contrast to policy gradient methods, which require workers to communicate entire gradients. 3) It does not require value function approximations. RL with value function estimation is inherently sequential: To improve upon a given policy, multiple updates to the value function are typically needed to get enough signal. Each time the policy is significantly changed, multiple iterations are necessary for the value function estimate to catch up.

A simple parallel version of ES is given in Algorithm 2. The main novelty here is that the algorithm makes use of shared random seeds, which drastically reduces the bandwidth required for communication between the workers.

**Algorithm 2** Parallelized Evolution Strategies

**Input:** Learning rate $\alpha$, noise standard deviation $\sigma$, initial policy parameters $\theta_0$
**Initialize:** $n$ workers with known random seeds, and initial parameters $\theta_0$
**for** $t = 0, 1, 2, \ldots$ **do**
  **for** each worker $i = 1, \ldots, n$ **do**
    Sample $\epsilon_i \sim \mathcal{N}(0, I)$
    Compute returns $F_i = F(\theta_t + \sigma\epsilon_i)$
  **end for**
  Send all scalar returns $F_i$ from each worker to every other worker
  **for** each worker $i = 1, \ldots, n$ **do**
    Reconstruct all perturbations $\epsilon_j$ for $j = 1, \ldots, n$ using known random seeds
    Set $\theta_{t+1} \leftarrow \theta_t + \alpha \frac{1}{n\sigma} \sum_{j=1}^{n} F_j \epsilon_j$
  **end for**
**end for**

In practice, we implement sampling by having each worker instantiate a large block of Gaussian noise at the start of training, and then perturbing its parameters by adding a randomly indexed subset of these noise variables at each iteration. Although this means that the perturbations are not strictly independent across iterations, we did not find this to be a problem in practice. Using this strategy, we find that the second part of Algorithm 2 (lines 9-12) only takes up a small fraction of total time spend for all our experiments, even when using up to 1,440 parallel workers. When using many more workers still, or when using very large neural networks, we can reduce the computation required for this part of the algorithm by having workers only perturb a subset of the parameters $\theta$ rather than all of them: In this case the perturbation distribution $p_\psi$ corresponds to a mixture of Gaussians, for which the update equations remain unchanged. At the very extreme, every worker would perturb only a single coordinate of the parameter vector, which means that we would be using pure finite differences.

To reduce variance, we use antithetic sampling Geweke [1988], also known as *mirrored sampling* Brockhoff et al. [2010] in the ES literature: that is, we always evaluate pairs of perturbations $\epsilon$, $-\epsilon$, for Gaussian noise vector $\epsilon$. We also find it useful to perform fitness shaping Wierstra et al. [2014] by applying a rank transformation to the returns before computing each parameter update. Doing so removes the influence of outlier individuals in each population and decreases the tendency for ES to fall into local optima early in training. In addition, we apply weight decay to the parameters of our policy network: this prevents the parameters from growing very large compared to the perturbations.

Unlike Wierstra et al. [2014] we did not see benefit from adapting $\sigma$ during training, and we therefore treat it as a fixed hyperparameter instead. We perform the optimization directly in parameter space; exploring indirect encodings Stanley et al. [2009], van Steenkiste et al. [2016] is left for future work.

Evolution Strategies, as presented above, works with full-length episodes. In some rare cases this can lead to low CPU utilization, as some episodes run for many more steps than others. For this reason, we cap episode length at a constant $m$ steps for all workers, which we dynamically adjust as training progresses. For example, by setting $m$ to be equal to twice the mean number of steps taken per episode, we can guarantee that CPU utilization stays above 50% in the worst case.

### 2.2 The impact of network parameterization

Whereas RL algorithms like Q-learning and policy gradients explore by sampling actions from a stochastic policy, Evolution Strategies derives learning signal from sampling instantiations of policy parameters. Exploration in ES is thus driven by parameter perturbation. For ES to improve upon parameters $\theta$, some members of the population must achieve better return than others: i.e. it is crucial that Gaussian perturbation vectors $\epsilon$ occasionally lead to new individuals $\theta + \sigma\epsilon$ with better return.

For the Atari environments, we found that Gaussian parameter perturbations on DeepMind's convolutional architectures [Mnih et al., 2015] did not always lead to adequate exploration: For some environments, randomly perturbed parameters tended to encode policies that always took one specific action regardless of the state that was given as input. However, we discovered thatwe could match the performance of policy gradient methods for most games by using virtual batch normalization [Salimans et al., 2016] in the policy specification. Virtual batch normalization is precisely equivalent to batch normalization [Ioffe and Szegedy, 2015] where the minibatch used for calculating normalizing statistics is chosen at the start of training and is fixed. This change in parameterization makes the policy more sensitive to very small changes in the input image at the early stages of training when the weights of the policy are random, ensuring that the policy takes a wide-enough variety of actions to gather occasional rewards. For most applications, a downside of virtual batch normalization is that it makes training more expensive. For our application, however, the minibatch used to calculate the normalizing statistics is much smaller than the number of steps taken during a typical episode, meaning that the overhead is negligible.

For the MuJoCo tasks, we achieved good performance on nearly all the environments with the standard multilayer perceptrons mapping to continuous actions. However, we observed that for some environments, we could encourage more exploration by discretizing the actions. This forced the actions to be non-smooth with respect to input observations and parameter perturbations, and thereby encouraged a wide variety of behaviors to be played out over the course of rollouts.

## 3 Smoothing in parameter space versus smoothing in action space

As mentioned in section 2, a large source of difficulty in RL stems from the lack of informative gradients of policy performance: such gradients may not exist due to non-smoothness of the environ-ment or policy, or may only be available as high-variance estimates because the environment usually can only be accessed via sampling. Explicitly, suppose we wish to solve general decision problems that give a return $R(\mathbf{a})$ after we take a sequence of actions $\mathbf{a} = \{a_1, \ldots, a_T\}$, where the actions are determined by a either a deterministic or a stochastic policy function $a_t = \pi(s; \theta)$. The objective we would like to optimize is thus

$$F(\theta) = R(\mathbf{a}(\theta)).$$

Since the actions are allowed to be discrete and the policy is allowed to be deterministic, $F(\theta)$ can be non-smooth in $\theta$. More importantly, because we do not have explicit access to the underlying state transition function of our decision problems, the gradients cannot be computed with a backpropagation-like algorithm. This means we cannot directly use standard gradient-based optimization methods to find a good solution for $\theta$.

In order to both make the problem smooth and to have a means of to estimate its gradients, we need to add noise. Policy gradient methods add the noise in action space, which is done by sampling the actions from an appropriate distribution. For example, if the actions are discrete and $\pi(s; \theta)$ calculates a score for each action before selecting the best one, then we would sample an action $\mathbf{a}(\epsilon, \theta)$ (here $\epsilon$ is the noise source) from a categorical distribution over actions at each time period, applying a softmax to the scores of each action. Doing so yields the objective $F_{PG}(\theta) = \mathbb{E}_\epsilon\, R(\mathbf{a}(\epsilon, \theta))$, with gradients

$$\nabla_\theta F_{PG}(\theta) = \mathbb{E}_\epsilon \left\{ R(\mathbf{a}(\epsilon, \theta)) \nabla_\theta \log p(\mathbf{a}(\epsilon, \theta); \theta) \right\}.$$

Evolution strategies, on the other hand, add the noise in parameter space. That is, they perturb the parameters as $\tilde{\theta} = \theta + \xi$, with $\xi$ from a multivariate Gaussian distribution, and then pick actions as $a_t = \mathbf{a}(\xi, \theta) = \pi(s; \tilde{\theta})$. It can be interpreted as adding a Gaussian blur to the original objective, which results in a smooth, differentiable cost $F_{ES}(\theta) = \mathbb{E}_\xi\, R(\mathbf{a}(\xi, \theta))$, this time with gradients

$$\nabla_\theta F_{ES}(\theta) = \mathbb{E}_\xi \left\{ R(\mathbf{a}(\xi, \theta)) \nabla_\theta \log p(\tilde{\theta}(\xi, \theta); \theta) \right\}.$$

The two methods for smoothing the decision problem are thus quite similar, and can be made even more so by adding noise to both the parameters and the actions.

### 3.1 When is ES better than policy gradients?

Given these two methods of smoothing the decision problem, which should we use? The answer depends strongly on the structure of the decision problem and on which type of Monte Carlo estimator is used to estimate the gradients $\nabla_\theta F_{PG}(\theta)$ and $\nabla_\theta F_{ES}(\theta)$. Suppose the correlation between the return and the individual actions is low (as is true for any hard RL problem). Assuming we approximate these gradients using simple Monte Carlo (REINFORCE) with a good baseline on the return, we have

$$\begin{aligned} \mathrm{Var}[\nabla_\theta F_{PG}(\theta)] &\approx \mathrm{Var}[R(\mathbf{a})]\, \mathrm{Var}[\nabla_\theta \log p(\mathbf{a}; \theta)], \\ \mathrm{Var}[\nabla_\theta F_{ES}(\theta)] &\approx \mathrm{Var}[R(\mathbf{a})]\, \mathrm{Var}[\nabla_\theta \log p(\tilde{\theta}; \theta)]. \end{aligned}$$

If both methods perform a similar amount of exploration, $\mathrm{Var}[R(\mathbf{a})]$ will be similar for both expressions. The difference will thus be in the second term. Here we have that $\nabla_\theta \log p(\mathbf{a}; \theta) = \sum_{t=1}^{T} \nabla_\theta \log p(a_t; \theta)$ is a sum of $T$ uncorrelated terms, so that the variance of the policy gradient estimator will grow nearly linearly with $T$. The corresponding term for evolution strategies, $\nabla_\theta \log p(\tilde{\theta}; \theta)$, is independent of $T$. Evolution strategies will thus have an advantage compared to policy gradients for long episodes with very many time steps. In practice, the effective number of steps $T$ is often reduced in policy gradient methods by discounting rewards. If the effects of actions are short-lasting, this allows us to dramatically reduce the variance in our gradient estimate, and this has been critical to the success of applications such as Atari games. However, this discounting will bias our gradient estimate if actions have long lasting effects. Another strategy for reducing the effective value of $T$ is to use value function approximation. This has also been effective, but once again runs the risk of biasing our gradient estimates. Evolution strategies is thus an attractive choice if the effective number of time steps $T$ is long, actions have long-lasting effects, and if no good value function estimates are available.

### 3.2 Problem dimensionality

The gradient estimate of ES can be interpreted as a method for randomized finite differences in high-dimensional space. Indeed, using the fact that $\mathbb{E}_{\epsilon\sim N(0,I)}\left\{F(\theta)\,\epsilon/\sigma\right\} = 0$, we get

$$\nabla_\theta \eta(\theta) = \mathbb{E}_{\epsilon\sim N(0,I)}\left\{F(\theta+\sigma\epsilon)\,\epsilon/\sigma\right\} \ = \mathbb{E}_{\epsilon\sim N(0,I)}\left\{(F(\theta+\sigma\epsilon) - F(\theta))\,\epsilon/\sigma\right\}$$

It is now apparent that ES can be seen as computing a finite difference derivative estimate in a randomly chosen direction, especially as $\sigma$ becomes small. The resemblance of ES to finite differences suggests the method will scale poorly with the dimension of the parameters $\theta$. Theoretical analysis indeed shows that for general non-smooth optimization problems, the required number of optimization steps scales linearly with the dimension [Nesterov and Spokoiny, 2011]. However, it is important to note that this does not mean that larger neural networks will perform worse than smaller networks when optimized using ES: what matters is the difficulty, or intrinsic dimension, of the optimization problem. To see that the dimensionality of our model can be completely separate from the effective dimension of the optimization problem, consider a regression problem where we approximate a univariate variable $y$ with a linear model $\hat{y} = \mathbf{x}\cdot\mathbf{w}$: if we double the number of features and parameters in this model by concatenating $\mathbf{x}$ with itself (i.e. using features $\mathbf{x}' = (\mathbf{x}, \mathbf{x})$), the problem does not become more difficult. The ES algorithm will do exactly the same thing when applied to this higher dimensional problem, as long as we divide the standard deviation of the noise by two, as well as the learning rate.

In practice, we observe slightly better results when using larger networks with ES. For example, we tried both the larger network and smaller network used in A3C [Mnih et al., 2016] for learning Atari 2600 games, and on average obtained better results using the larger network. We hypothesize that this is due to the same effect that makes standard gradient-based optimization of large neural networks easier than for small ones: large networks have fewer local minima [Kawaguchi, 2016].

### 3.3 Advantages of not calculating gradients

In addition to being easy to parallelize, and to having an advantage in cases with long action sequences and delayed rewards, black box optimization algorithms like ES have other advantages over RL techniques that calculate gradients. The communication overhead of implementing ES in a distributed setting is lower than for competing RL methods such as policy gradients and Q-learning, as the only information that needs to be communicated across processes are the scalar return and the random seed that was used to generate the perturbations $\epsilon$, rather than a full gradient. Also, ES can deal with maximally sparse and delayed rewards; only the total return of an episode is used, whereas other methods use individual rewards and their exact timing.

By not requiring backpropagation, black box optimizers reduce the amount of computation per episode by about two thirds, and memory by potentially much more. In addition, not explicitly calculating an analytical gradient protects against problems with exploding gradients that are common when working with recurrent neural networks. By smoothing the cost function in parameter space, we reduce the pathological curvature that causes these problems: bounded cost functions that are smooth enough can't have exploding gradients. At the extreme, ES allows us to incorporate non-differentiable elements into our architecture, such as modules that use *hard attention* [Xu et al., 2015].

Black box optimization methods are uniquely suited to low precision hardware for deep learning. Low precision arithmetic, such as in binary neural networks, can be performed much cheaper than at high precision. When optimizing such low precision architectures, biased low precision gradient estimates can be a problem when using gradient-based methods. Similarly, specialized hardware for neural network inference, such as TPUs [Jouppi et al., 2017], can be used directly when performing optimization using ES, while their limited memory usually makes backpropagation impossible.

By perturbing in parameter space instead of action space, black box optimizers are naturally invariant to the frequency at which our agent acts in the environment. For MDP-based reinforcement learning algorithms, on the other hand, it is well known that *frameskip* is a crucial parameter to get right for the optimization to succeed [Braylan et al., 2005]. While this is usually a solvable problem for games that only require short-term planning and action, it is a problem for learning longer term strategic behavior. For these problems, RL needs hierarchy to succeed [Parr and Russell, 1998], which is not as necessary when using black box optimization.

## 4 Experiments

### 4.1 MuJoCo

We evaluated ES on a benchmark of continuous robotic control problems in the OpenAI Gym [Brockman et al., 2016] against a highly tuned implementation of Trust Region Policy Optimization [Schulman et al., 2015], a policy gradient algorithm designed to efficiently optimize neural network policies. We tested on both classic problems, like balancing an inverted pendulum, and more difficult ones found in recent literature, like learning 2D hopping and walking gaits. The environments were simulated by MuJoCo [Todorov et al., 2012].

We used both ES and TRPO to train policies with identical architectures: multilayer perceptrons with two 64-unit hidden layers separated by tanh nonlinearities. We found that ES occasionally benefited from discrete actions, since continuous actions could be too smooth with respect to parameter perturbation and could hamper exploration (see section 2.2). For the hopping and swimming tasks, we discretized the actions for ES into 10 bins for each action component.

We found that ES was able to solve these tasks up to TRPO's final performance after 5 million timesteps of environment interaction. To obtain this result, we ran ES over 6 random seeds and compared the mean learning curves to similarly computed curves for TRPO. The exact sample complexity tradeoffs over the course of learning are listed in Table 1, and detailed results are listed in Table 3 of the supplement. Generally, we were able to solve the environments in less than 10x penalty in sample complexity on the hard environments (Hopper and Walker2d) compared to TRPO. On simple environments, we achieved up to 3x better sample complexity than TRPO.

Table 1: MuJoCo tasks: Ratio of ES timesteps to TRPO timesteps needed to reach various percentages of TRPO's learning progress at 5 million timesteps.

| Environment | 25% | 50% | 75% | 100% |
|---|---|---|---|---|
| HalfCheetah | 0.15 | 0.49 | 0.42 | 0.58 |
| Hopper | 0.53 | 3.64 | 6.05 | 6.94 |
| InvertedDoublePendulum | 0.46 | 0.48 | 0.49 | 1.23 |
| InvertedPendulum | 0.28 | 0.52 | 0.78 | 0.88 |
| Swimmer | 0.56 | 0.47 | 0.53 | 0.30 |
| Walker2d | 0.41 | 5.69 | 8.02 | 7.88 |

### 4.2 Atari

We ran our parallel implementation of Evolution Strategies, described in Algorithm 2, on 51 Atari 2600 games available in OpenAI Gym [Brockman et al., 2016]. We used the same preprocessing and feedforward CNN architecture used by Mnih et al. [2016]. All games were trained for 1 billion frames, which requires about the same amount of neural network computation as the published 1-day results for A3C [Mnih et al., 2016] which uses 320 million frames. The difference is due to the fact that ES does not perform backpropagation and does not use a value function. By parallelizing the evaluation of perturbed parameters across 720 CPUs on Amazon EC2, we can bring down the time required for the training process to about one hour per game. After training, we compared final performance against the published A3C results and found that ES performed better in 23 games tested, while it performed worse in 28. The full results are in Table 2 in the supplementary material.

### 4.3 Parallelization

ES is particularly amenable to parallelization because of its low communication bandwidth requirement (Section 2.1). We implemented a distributed version of Algorithm 2 to investigate how ES scales with the number of workers. Our distributed implementation did not rely on special networking setup and was tested on public cloud computing service Amazon EC2.

We picked the 3D Humanoid walking task from OpenAI Gym [Brockman et al., 2016] as the test problem for our scaling experiment, because it is one of the most challenging continuous control problems solvable by state-of-the-art RL techniques, which require about a day to learn on modern hardware [Schulman et al., 2015, Duan et al., 2016a]. Solving 3D Humanoid with ES on one 18-core machine takes about 11 hours, which is on par with RL. However, when distributed across 80 machines and $1,440$ CPU cores, ES can solve 3D Humanoid in just 10 minutes, reducing experiment turnaround time by two orders of magnitude. Figure 1 shows that, for this task, ES is able to achieve linear speedup in the number of CPU cores.

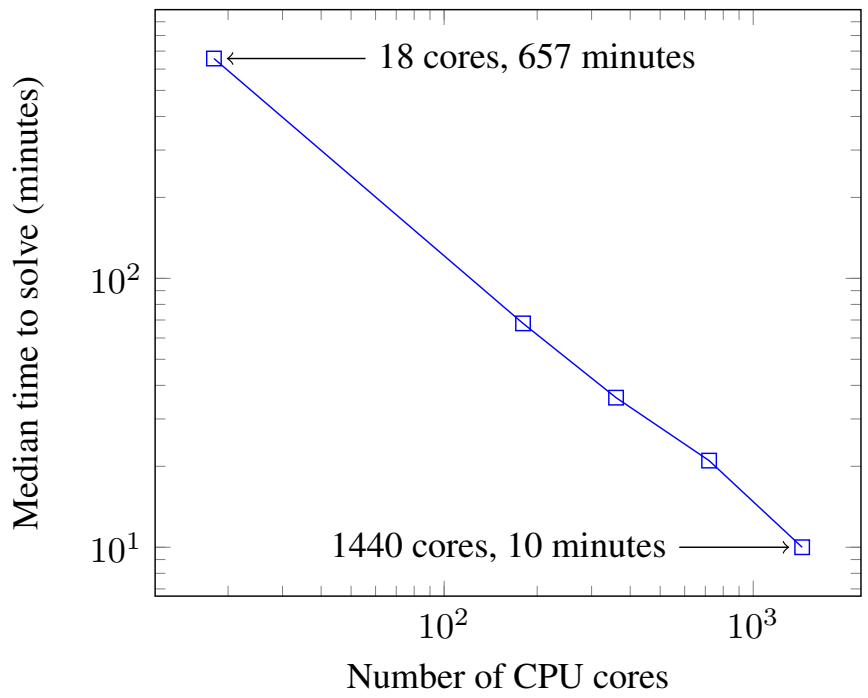

Figure 1: Time to reach a score of 6000 on 3D Humanoid with different number of CPU cores. Experiments are repeated 7 times and median time is reported.

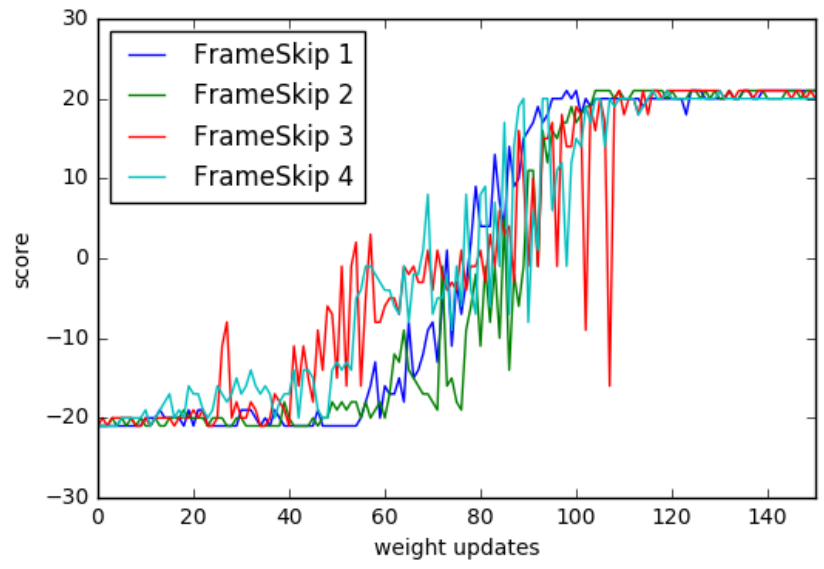

Figure 2: Learning curves for Pong using varying frame-skip parameters. Although performance is stochastic, each setting leads to about equally fast learning, with each run converging in around 100 weight updates.

### 4.4 Invariance to temporal resolution

It is common practice in RL to have the agent decide on its actions in a lower frequency than is used in the simulator that runs the environment. This action frequency, or *frame-skip*, is a crucial parameter in many RL algorithms [Braylan et al., 2005]. If the frame-skip is set too high, the agent cannot make its decisions at a fine enough timeframe to perform well in the environment. If, on the other hand, the frameskip is set too low, the effective time length of the episode increases too much, which deteriorates optimization performance as analyzed in section 3.1. An advantage of ES is that its gradient estimate is invariant to the length of the episode, which makes it much more robust to the action frequency. We demonstrate this by running the Atari game Pong using a frame skip parameter in $\{1, 2, 3, 4\}$. As can be seen in Figure 2, the learning curves for each setting indeed look very similar.

## 5 Related work

There have been many attempts at applying methods related to ES to train neural networks Risi and Togelius [2015]. For Atari, Hausknecht et al. [2014] obtain impressive results. Sehnke et al. [2010] proposed a method closely related the one investigated in our work. Koutník et al. [2013, 2010] and Srivastava et al. [2012] have similarly applied an an ES method to RL problems with visual inputs, but where the policy was compressed in a number of different ways. Natural evolution strategies has been successfully applied to black box optimization Wierstra et al. [2008, 2014], as well as for the training of the recurrent weights in recurrent neural networks Schmidhuber et al. [2007]. Stulp and Sigaud [2012] explored similar approaches to black box optimization. An interesting hybrid of black-box optimization and policy gradient methods was recently explored by Usunier et al. [2016]. Hyper-Neat Stanley et al. [2009] is an alternative approach to evolving both the weights of the neural networks and their parameters. Derivative free optimization methods have also been analyzed in the convex setting Duchi et al. [2015], Nesterov [2012].

The main contribution in our work is in showing that this class of algorithms is extremely scalable and efficient to use on distributed hardware. We have shown that ES, when carefully implemented, is competitive with competing RL algorithms in terms of performance on the hardest problems solvable today, and is surprisingly close in terms of data efficiency, while taking less wallclock time to train.

## 6 Conclusion

We have explored Evolution Strategies, a class of black-box optimization algorithms, as an alternative to popular MDP-based RL techniques such as Q-learning and policy gradients. Experiments on Atari and MuJoCo show that it is a viable option with some attractive features: it is invariant to action frequency and delayed rewards, and it does not need temporal discounting or value function approximation. Most importantly, ES is highly parallelizable, which allows us to make up for a decreased data efficiency by scaling to more parallel workers.

In future work, we plan to apply evolution strategies to those problems for which MDP-based reinforcement learning is less well-suited: problems with long time horizons and complicated reward structure. We are particularly interested in meta-learning, or learning-to-learn. A proof of concept for meta-learning in an RL setting was given by Duan et al. [2016b]: Using black-box optimization we hope to be able to extend these results. We also plan to examine combining ES with fast low precision neural network implementations to fully make use of the gradient-free nature of ES.

| Game | DQN | A3C FF, 1 day | HyperNEAT | ES FF, 1 hour | A2C FF |
|---|---|---|---|---|---|
| Amidar | 133.4 | 283.9 | 184.4 | 112.0 | **548.2** |
| Assault | 3332.3 | **3746.1** | 912.6 | 1673.9 | 2026.6 |
| Asterix | 124.5 | **6723.0** | 2340.0 | 1440.0 | 3779.7 |
| Asteroids | 697.1 | **3009.4** | 1694.0 | 1562.0 | 1733.4 |
| Atlantis | 76108.0 | 772392.0 | 61260.0 | 1267410.0 | **2872644.8** |
| Bank Heist | 176.3 | **946.0** | 214.0 | 225.0 | 724.1 |
| Battle Zone | 17560.0 | 11340.0 | **36200.0** | 16600.0 | 8406.2 |
| Beam Rider | 8672.4 | **13235.9** | 1412.8 | 744.0 | 4438.9 |
| Berzerk | | **1433.4** | 1394.0 | 686.0 | 720.6 |
| Bowling | 41.2 | 36.2 | **135.8** | 30.0 | 28.9 |
| Boxing | 25.8 | 33.7 | 16.4 | 49.8 | **95.8** |
| Breakout | 303.9 | **551.6** | 2.8 | 9.5 | 368.5 |
| Centipede | 3773.1 | 3306.5 | **25275.2** | 7783.9 | 2773.3 |
| Chopper Command | 3046.0 | **4669.0** | 3960.0 | 3710.0 | 1700.0 |
| Crazy Climber | 50992.0 | **101624.0** | 0.0 | 26430.0 | 100034.4 |
| Demon Attack | 12835.2 | **84997.5** | 14620.0 | 1166.5 | 23657.7 |
| Double Dunk | **21.6** | 0.1 | 2.0 | 0.2 | 3.2 |
| Enduro | **475.6** | 82.2 | 93.6 | 95.0 | 0.0 |
| Fishing Derby | 2.3 | 13.6 | **49.8** | 49.0 | 33.9 |
| Freeway | 25.8 | 0.1 | 29.0 | **31.0** | 0.0 |
| Frostbite | 157.4 | 180.1 | **2260.0** | 370.0 | 266.6 |
| Gopher | 2731.8 | **8442.8** | 364.0 | 582.0 | 6266.2 |
| Gravitar | 216.5 | 269.5 | 370.0 | **805.0** | 256.2 |
| Ice Hockey | 3.8 | 4.7 | **10.6** | 4.1 | 4.9 |
| Kangaroo | 2696.0 | 106.0 | 800.0 | **11200.0** | 1357.6 |
| Krull | 3864.0 | 8066.6 | **12601.4** | 8647.2 | 6411.5 |
| Montezuma's Revenge | 50.0 | **53.0** | 0.0 | 0.0 | 0.0 |
| Name This Game | 5439.9 | 5614.0 | **6742.0** | 4503.0 | 5532.8 |
| Phoenix | | **28181.8** | 1762.0 | 4041.0 | 14104.7 |
| Pit Fall | | **123.0** | 0.0 | 0.0 | 8.2 |
| Pong | 16.2 | 11.4 | 17.4 | **21.0** | 20.8 |
| Private Eye | 298.2 | 194.4 | **10747.4** | 100.0 | 100.0 |
| Q*Bert | 4589.8 | 13752.3 | 695.0 | 147.5 | **15758.6** |
| River Raid | 4065.3 | **10001.2** | 2616.0 | 5009.0 | 9856.9 |
| Road Runner | 9264.0 | 31769.0 | 3220.0 | 16590.0 | **33846.9** |
| Robotank | **58.5** | 2.3 | 43.8 | 11.9 | 2.2 |
| Seaquest | **2793.9** | 2300.2 | 716.0 | 1390.0 | 1763.7 |
| Skiing | | 13700.0 | 7983.6 | **15442.5** | 15245.8 |
| Solaris | | 1884.8 | 160.0 | 2090.0 | **2265.0** |
| Space Invaders | 1449.7 | **2214.7** | 1251.0 | 678.5 | 951.9 |
| Star Gunner | 34081.0 | **64393.0** | 2720.0 | 1470.0 | 40065.6 |
| Tennis | 2.3 | 10.2 | 0.0 | 4.5 | **11.2** |
| Time Pilot | 5640.0 | 5825.0 | **7340.0** | 4970.0 | 4637.5 |
| Tutankham | 32.4 | 26.1 | 23.6 | 130.3 | **194.3** |
| Up and Down | 3311.3 | 54525.4 | 43734.0 | 67974.0 | **75785.9** |
| Venture | 54.0 | 19.0 | 0.0 | **760.0** | 0.0 |
| Video Pinball | 20228.1 | **185852.6** | 0.0 | 22834.8 | 46470.1 |
| Wizard of Wor | 246.0 | **5278.0** | 3360.0 | 3480.0 | 1587.5 |
| Yars Revenge | | 7270.8 | **24096.4** | 16401.7 | 8963.5 |
| Zaxxon | 831.0 | 2659.0 | 3000.0 | **6380.0** | 5.6 |

Table 2: Final results obtained using Evolution Strategies on Atari 2600 games (feedforward CNN policy, deterministic policy evaluation, averaged over 10 re-runs with up to 30 random initial no-ops), and compared to results for DQN and A3C from Mnih et al. [2016] and HyperNEAT from Hausknecht et al. [2014]. A2C is our synchronous variant of A3C, and its reported scores are obtained with 320M training frames with the same evaluation setup as for the ES results. All methods were trained on raw pixel input.

Table 3: MuJoCo tasks: Ratio of ES timesteps to TRPO timesteps needed to reach various percentages of TRPO's learning progress at 5 million timesteps. These results were computed from ES learning curves averaged over 6 reruns.

| Environment | % TRPO final score | TRPO score | TRPO timesteps | ES timesteps | ES timesteps / TRPO timesteps |
|---|---|---|---|---|---|
| HalfCheetah | 25% | -1.35 | 9.05e+05 | 1.36e+05 | 0.15 |
| | 50% | 793.55 | 1.70e+06 | 8.28e+05 | 0.49 |
| | 75% | 1589.83 | 3.34e+06 | 1.42e+06 | 0.42 |
| | 100% | 2385.79 | 5.00e+06 | 2.88e+06 | 0.58 |
| Hopper | 25% | 877.45 | 7.29e+05 | 3.83e+05 | 0.53 |
| | 50% | 1718.16 | 1.03e+06 | 3.73e+06 | 3.64 |
| | 75% | 2561.11 | 1.59e+06 | 9.63e+06 | 6.05 |
| | 100% | 3403.46 | 4.56e+06 | 3.16e+07 | 6.94 |
| InvertedDoublePendulum | 25% | 2358.98 | 8.73e+05 | 3.98e+05 | 0.46 |
| | 50% | 4609.68 | 9.65e+05 | 4.66e+05 | 0.48 |
| | 75% | 6874.03 | 1.07e+06 | 5.30e+05 | 0.49 |
| | 100% | 9104.07 | 4.39e+06 | 5.39e+06 | 1.23 |
| InvertedPendulum | 25% | 276.59 | 2.21e+05 | 6.25e+04 | 0.28 |
| | 50% | 519.15 | 2.73e+05 | 1.43e+05 | 0.52 |
| | 75% | 753.17 | 3.25e+05 | 2.55e+05 | 0.78 |
| | 100% | 1000.00 | 5.17e+05 | 4.55e+05 | 0.88 |
| Swimmer | 25% | 41.97 | 1.04e+06 | 5.88e+05 | 0.56 |
| | 50% | 70.73 | 1.82e+06 | 8.52e+05 | 0.47 |
| | 75% | 99.68 | 2.33e+06 | 1.23e+06 | 0.53 |
| | 100% | 128.25 | 4.59e+06 | 1.39e+06 | 0.30 |
| Walker2d | 25% | 957.68 | 1.55e+06 | 6.43e+05 | 0.41 |
| | 50% | 1916.48 | 2.27e+06 | 1.29e+07 | 5.69 |
| | 75% | 2872.81 | 2.89e+06 | 2.31e+07 | 8.02 |
| | 100% | 3830.03 | 4.81e+06 | 3.79e+07 | 7.88 |